\documentclass[conference]{IEEEtran}
\IEEEoverridecommandlockouts
\usepackage{blindtext}
\usepackage{hyperref}
\usepackage{amsmath,amssymb,amsfonts}
\usepackage{caption}
\usepackage{subcaption}
\usepackage{graphicx}
\usepackage{textcomp}
\usepackage{xcolor}
\usepackage{url}
\usepackage{booktabs}
\usepackage{algorithm}
\usepackage{algpseudocode}
\usepackage{amsfonts}
\usepackage{tikz}

\newcommand{\ndim}{d}                
\newcommand{\moff}{m}                
\newcommand{\rpop}{\rho}                 
\newcommand{\rhat}{\widehat{\rho}}       
\newcommand{\rhatplus}{\widehat{\rho}_+} 

\usetikzlibrary{positioning,arrows.meta,calc,fit,decorations.pathreplacing}
\tikzset{in/.style={circle, draw, minimum size=6mm}}
\def\BibTeX{{\rm B\kern-.05em{\sc i\kern-.025em b}\kern-.08em
    T\kern-.1667em\lower.7ex\hbox{E}\kern-.125emX}}
\begin{document}

\title{Evaluating the Efficiency of Latent Spaces\\via the Coupling-Matrix}

\author{
  \IEEEauthorblockN{Mehmet Can Yavuz}
  \IEEEauthorblockA{
    \textit{Faculty of Engineering and Natural Sciences} \\
    Işık University, İstanbul, Türkiye \\
    mehmetcan.yavuz@isikun.edu.tr
  }
  \and
  \IEEEauthorblockN{Berrin Yanikoglu}
  \IEEEauthorblockA{
    \textit{Faculty of Engineering and Natural Sciences} \\
    Sabancı University, İstanbul, Türkiye \\
    berrin@sabanciuniv.edu
  }
}

\maketitle

\begin{abstract}
A central challenge in representation learning is constructing latent embeddings that are both expressive and efficient. In practice, deep networks often produce redundant latent spaces where multiple coordinates encode overlapping information, reducing effective capacity and hindering generalization. Standard metrics such as accuracy or reconstruction loss provide only indirect evidence of such redundancy and cannot isolate it as a failure mode. We introduce a redundancy index, denoted $\rho(C)$, that directly quantifies inter-dimensional dependencies by analyzing coupling matrices derived from latent representations and comparing their off-diagonal statistics against a normal distribution via energy distance. The result is a compact, interpretable, and statistically grounded measure of representational quality. We validate $\rho(C)$ across discriminative and generative settings on MNIST variants, Fashion-MNIST, CIFAR-10, and CIFAR-100, spanning multiple architectures and hyperparameter optimization strategies. Empirically, low $\rho(C)$ reliably predicts high classification accuracy or low reconstruction error, while elevated redundancy is associated with performance collapse. Estimator reliability grows with latent dimension, yielding natural lower bounds for reliable analysis. We further show that Tree-structured Parzen Estimators (TPE) preferentially explore low-$\rho$ regions, suggesting that $\rho(C)$ can guide neural architecture search and serve as a redundancy-aware regularization target. By exposing redundancy as a universal bottleneck across models and tasks, $\rho(C)$ offers both a theoretical lens and a practical tool for evaluating and improving the efficiency of learned representations.
\end{abstract}


\section{Introduction}

A persistent challenge in representation learning is to construct latent embeddings that are simultaneously expressive and efficient. Deep neural networks, while highly flexible, often produce redundant representations in which multiple latent coordinates encode overlapping information. Such redundancy wastes representational capacity and can degrade generalization to unseen data. Yet, despite its importance, redundancy remains difficult to measure directly. Standard metrics such as classification accuracy or reconstruction loss offer only indirect evidence of representational quality and cannot disentangle redundancy from other sources of error.

The importance of disentanglement is widely acknowledged in both discriminative and generative modeling. From an information-theoretic perspective, an embedding achieves maximal capacity when its dimensions are statistically independent, ensuring that each channel contributes novel information. Correlated or redundant dimensions reduce joint entropy and limit the embedding’s ability to support downstream tasks. Existing approaches to measuring such dependencies—such as mutual information estimation—are computationally demanding and unstable in high dimensions, leaving a gap between theory and practice.

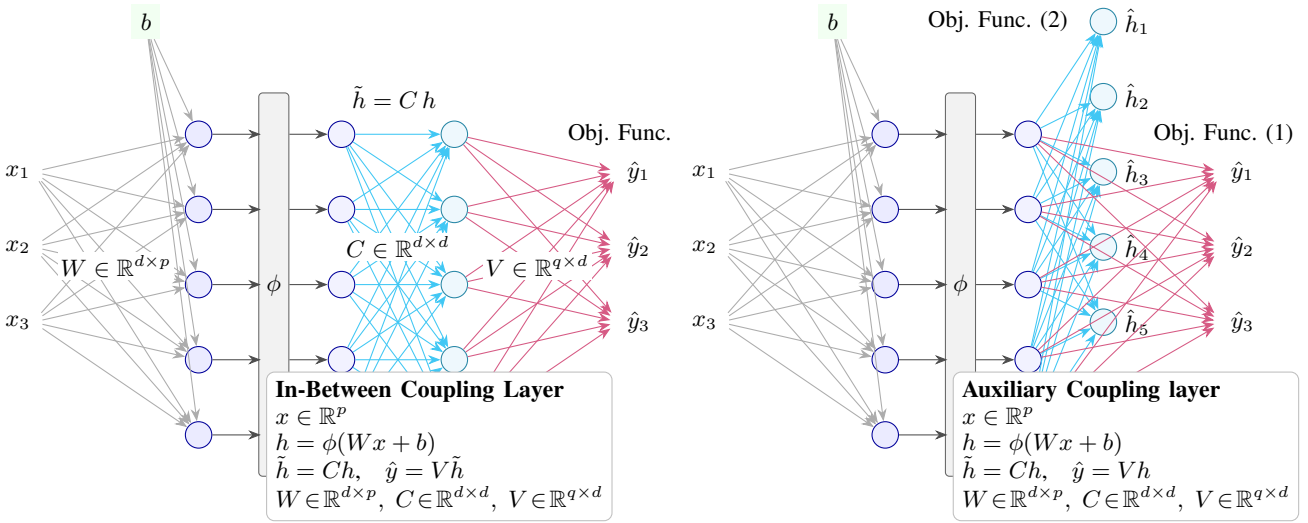
\begin{figure*}[htb!]
  \centering
  \begin{tikzpicture}[
      >=Stealth,
      node distance=1.2cm and 2.0cm,
      on grid,
      every node/.style={font=\small},
      in/.style={circle, draw=green!60!black, fill=green!10, minimum size=10pt},
      hid/.style={circle, draw=blue!60!black, fill=blue!8, minimum size=10pt},
      out/.style={circle, draw=purple!60!black, fill=purple!8, minimum size=10pt},
      op/.style={rectangle, rounded corners=1.5pt, draw=black!60, fill=gray!10, inner sep=2pt},
      matlabel/.style={fill=white, inner sep=1pt}
    ]
    \def\ninput{3} 
    \def\nhidden{5} 
    \def\noutput{3} 
    \def\xshift{0}
    \def\yshift{0}
    \def\xgapA{2.2} 
    \def\xgapB{1.0} 
    \def\xgapC{2.4} 
    \def\xgapD{2.2} 
    \foreach \i in {1,...,\ninput} {
      \node (x\i) at (\xshift, \yshift - \i) {};
      \node[left=0.2cm of x\i] {$x_{\i}$};
      }
    \node[fill=green!5] (b) at (\xshift + 1.5, \yshift - \ninput + 4) {$b$};
    \foreach \j in {1,...,\nhidden} {
      \node[hid] (y\j) at (\xshift+\xgapA, \yshift - \j + 0.5) {};
    }
    \foreach \i in {1,...,\ninput} {
      \foreach \j in {1,...,\nhidden} {
        \draw[->, gray!65] (x\i) -- (y\j);
      }
    }
    \foreach \j in {1,...,\nhidden} {
      \draw[->, gray!65] (b) -- (y\j);
    }
    \node[matlabel] at ($ (x2)!0.5!(y3) + (0.0,0) $) {$W \in \mathbb{R}^{d \times p}$};
    
    \node[op, minimum width=0.4cm, minimum height=5.1cm, inner sep=0pt]
      (phi) at ($ (y1)!0.5!(y\nhidden) + (\xgapB,0) $) {$\phi$};
    
    \foreach \j in {1,...,\nhidden} {
      \draw[->, black!70, shorten >= -1pt] (y\j) -- (phi.west |- y\j);
    }
    
    \foreach \j in {1,...,\nhidden} {
      \node[hid, fill=blue!5] (h\j) at (\xshift+\xgapA+\xgapB+0.9, \yshift - \j + 0.5) {};
      \draw[->, black!70] (phi.east|-y\j) -- (h\j);
    }
    \foreach \k in {1,...,\nhidden} {
      \node[hid, fill=cyan!6, draw=cyan!60!black] (t\k) at (\xshift+\xgapA+\xgapB+\xgapC, \yshift - \k + 0.5) {};
    }
    \foreach \j in {1,...,\nhidden} {
      \foreach \k in {1,...,\nhidden} {
        \draw[->, cyan!60] (h\j) -- (t\k);
      }
    }
    \node[matlabel] at ($ (h2)!0.5!(t3) $) {$C \in \mathbb{R}^{d \times d}$};
    \node[matlabel, above=0.5cm of t1, xshift=-0.8cm] {$\tilde{h}=C\,h$};
    \foreach \o in {1,...,\noutput} {
      \node (o\o) at (\xshift+\xgapA+\xgapB+\xgapC+\xgapD, \yshift - \o) {}; 
      \node[right=0.25cm of o\o] {$\hat{y}_{\o}$};
    }
    \foreach \k in {1,...,\nhidden} {
      \foreach \o in {1,...,\noutput} {
        \draw[->, purple!65] (t\k) -- (o\o);
      }
    }
    \node[matlabel, above=0.5cm of o1, xshift=0.0cm] {Obj. Func.};
    \node[draw=black!25, rounded corners=3pt, fill=white, align=left, anchor=north]
      at ($ (phi.south) + (2.2,1.4) $) {%
        \textbf{In-Between Coupling Layer}\\
        $x \in \mathbb{R}^{p}$\\
        $h=\phi(Wx+b)$\\
        $\tilde{h} = C h$,\quad $\hat{y} = V\tilde{h}$\\[1pt]
        $W\!\in\!\mathbb{R}^{d\times p},\; C\!\in\!\mathbb{R}^{d\times d},\; V\!\in\!\mathbb{R}^{q\times d}$
      };    
    \node[matlabel] at ($ (t3)!0.5!(o2) $) {$V \in \mathbb{R}^{q \times d}$};
  \end{tikzpicture}
    \begin{tikzpicture}[
      >=Stealth,
      node distance=1.2cm and 2.0cm,
      on grid,
      every node/.style={font=\small},
      in/.style={circle, draw=green!60!black, fill=green!10, minimum size=10pt},
      hid/.style={circle, draw=blue!60!black, fill=blue!8, minimum size=10pt},
      out/.style={circle, draw=purple!60!black, fill=purple!8, minimum size=10pt},
      op/.style={rectangle, rounded corners=1.5pt, draw=black!60, fill=gray!10, inner sep=2pt},
      matlabel/.style={fill=white, inner sep=1pt}
    ]
    \def\ninput{3} 
    \def\nhidden{5} 
    \def\noutput{3} 
    \def\xshift{0}
    \def\yshift{0}
    \def\xgapA{2.2} 
    \def\xgapB{1.0} 
    \def\xgapC{2.4} 
    \def\xgapD{2.2} 
    \foreach \i in {1,...,\ninput} {
      \node (x\i) at (\xshift, \yshift - \i) {};
      \node[left=0.2cm of x\i] {$x_{\i}$};
      }
    \node[fill=green!5] (b) at (\xshift + 1.5, \yshift - \ninput + 4) {$b$};
    \foreach \j in {1,...,\nhidden} {
      \node[hid] (y\j) at (\xshift+\xgapA, \yshift - \j + 0.5) {};
    }
    \foreach \i in {1,...,\ninput} {
      \foreach \j in {1,...,\nhidden} {
        \draw[->, gray!65] (x\i) -- (y\j);
      }
    }
    \foreach \j in {1,...,\nhidden} {
      \draw[->, gray!65] (b) -- (y\j);
    }
    
    \node[op, minimum width=0.4cm, minimum height=5.1cm, inner sep=0pt]
      (phi) at ($ (y1)!0.5!(y\nhidden) + (\xgapB,0) $) {$\phi$};
    
    \foreach \j in {1,...,\nhidden} {
      \draw[->, black!70, shorten >= -1pt] (y\j) -- (phi.west |- y\j);
    }
    
    \foreach \j in {1,...,\nhidden} {
      \node[hid, fill=blue!5] (h\j) at (\xshift+\xgapA+\xgapB+0.9, \yshift - \j + 0.5) {};
      \draw[->, black!70] (phi.east|-y\j) -- (h\j);
    }
    \foreach \k in {1,...,\nhidden} {
      \node[hid, fill=cyan!6, draw=cyan!60!black] (t\k) at (\xshift+\xgapA+\xgapB+\xgapC-0.5, \yshift - \k + 2.0) {};
      \node[right=0.45cm of t\k] {$\hat{h}_{\k}$};
    }
    \foreach \j in {1,...,\nhidden} {
      \foreach \k in {1,...,\nhidden} {
        \draw[->, cyan!60] (h\j) -- (t\k);
      }
    }
    \node[matlabel, above=0.0cm of t1, xshift=-1.4cm] {Obj. Func. (2)};
    \foreach \o in {1,...,\noutput} {
      \node (o\o) at (\xshift+\xgapA+\xgapB+\xgapC+\xgapD/2, \yshift - \o) {}; 
      \node[right=0.25cm of o\o] {$\hat{y}_{\o}$};
    }
    \foreach \k in {1,...,\nhidden} {
      \foreach \o in {1,...,\noutput} {
        \draw[->, purple!65] (h\k) -- (o\o);
      }
    }
    \node[matlabel, above=0.5cm of o1, xshift=0.0cm] {Obj. Func. (1)};
    \node[draw=black!25, rounded corners=3pt, fill=white, align=left, anchor=north]
      at ($ (phi.south) + (2.2,1.4) $) {%
        \textbf{Auxiliary Coupling layer}\\
        $x \in \mathbb{R}^{p}$\\
        $h=\phi(Wx+b)$\\
        $\tilde{h} = C h$,\quad $\hat{y} = Vh$\\[1pt]
        $W\!\in\!\mathbb{R}^{d\times p},\; C\!\in\!\mathbb{R}^{d\times d},\; V\!\in\!\mathbb{R}^{q\times d}$
      };    
  \end{tikzpicture}  
  \caption{Comparison of coupling strategies in a one-layer MLP. 
\textbf{(Left)} In-between coupling: the hidden representation \(h\) is first transformed 
by an internal coupling layer \(C\) before projection to the outputs, yielding the pipeline 
\(x \xrightarrow{W,b} h \xrightarrow{C} \tilde{h} \xrightarrow{V} \hat{y}\). 
\textbf{(Right)} Auxiliary coupling: the hidden representation \(h\) branches into two parallel 
paths, one directly projected to the outputs \(h \xrightarrow{V} \hat{y}\), 
and another transformed by \(C\) into \(\tilde{h}\) for auxiliary analysis.}
  \label{fig:coupling_strategies}
\end{figure*}

We address this gap by introducing a redundancy index, denoted $\rho(C)$, which provides a simple yet principled measurement of representational quality. Fig.\ref{fig:coupling_strategies} illustrates how coupling layers are incorporated into a one-layer MLP, comparing in-between and auxiliary coupling strategies that serve as the basis for defining and analyzing the redundancy index~$\rho(C)$. By comparing the distribution of off-diagonal coupling coefficients to a normal distribution model using the energy distance, $\rho(C)$ directly quantifies redundancy in a compact, interpretable, and statistically grounded way. In an ideally disentangled regime, off-diagonal entries resemble Gaussian noise; systematic deviations signal redundancy.

We validate $\rho(C)$ through extensive experiments spanning both classification and generative modeling. Using MNIST variants, Fashion-MNIST, CIFAR-10, and CIFAR-100, we examine $\rho(C)$ across multiple neural architectures and hyperparameter optimization strategies. Our analyses address the following research questions:

\begin{itemize}
    \item[\textbf{Q1}] How does $\rho(C)$ correlate with supervised accuracy in discriminative models?
    \item[\textbf{Q2}] Can $\rho(C)$ estimation analyses establish reliable lower bounds on latent dimensionality?
    \item[\textbf{Q3}] Can $\rho(C)$ capture representational dynamics during training and highlight promising hyperparameter configurations?
    \item[\textbf{Q4}] Does the $\rho$–performance relationship extend consistently to deeper architectures?
    \item[\textbf{Q5}] Does $\rho(C)$ generalize as a measurement for generative models such as autoencoders?
\end{itemize}

Our results demonstrate that $\rho(C)$ is predictive, stable, and generalizable. $\rhat$ variance decreases with latent dimension with variance shrinking as the effective number of off-diagonal samples grows (see Exp. 2), and low $\rho$ consistently predicts strong performance across tasks. We also find that search strategies such as Tree-structured Parzen Estimators (TPE) naturally favor low-$\rho$ regions, suggesting that $\rho$ can inform architecture design and optimization. Finally, we show that redundancy thresholds apply consistently across discriminative and generative models, reinforcing the universality of $\rho(C)$ as a measurement of representational quality.

\textbf{Contributions.}  
This paper makes the following contributions:
\begin{enumerate}
    \item We propose a theoretically grounded redundancy index $\rho(C)$, derived from energy distance comparisons of coupling matrices, as a compact measurement of latent space quality.
    \item We provide the first systematic study linking $\rho(C)$ to generalization across diverse datasets, architectures, and training strategies, identifying consistent thresholds that separate robust from redundant embeddings.
    \item We show that $\rho(C)$ reveals architectural trade-offs—such as redundancy explosions in overly wide layers—that are obscured by accuracy alone.
    \item We demonstrate the generality of $\rho(C)$ by validating its predictive role in both discriminative and generative models, establishing redundancy as a universal bottleneck in representation learning.
\end{enumerate}

By grounding redundancy measurement in statistical discrepancy theory and validating it empirically across tasks, this work provides both a measurement tool and a conceptual framework for understanding and improving the efficiency of learned representations.

\section{Related Works}

\subsection{Disentanglement and Redundancy in Representations}
A longstanding goal of representation learning is to discover latent spaces in which factors of variation are disentangled and non-redundant. Early frameworks such as the $\beta$-VAE \cite{higgins2017beta} and FactorVAE \cite{kim2018disentangling} formalized disentanglement as the statistical independence of latent variables. These works highlighted the connection between redundancy and degraded generalization, showing that correlated embeddings hinder transfer across tasks. More recent studies \cite{locatello2019challenging, eastwood2018framework} emphasize that perfect disentanglement is neither identifiable nor always necessary, motivating instead measurement tools for quantifying redundancy and correlation.

\subsection{Information-Theoretic Quality Measures}
Information-theoretic approaches provide a natural foundation for evaluating latent spaces. Mutual information has been widely used to assess representational capacity \cite{belghazi2018mutual, tschannen2019mutual}, but its estimation is notoriously difficult in high dimensions. Alternative measures such as Total Correlation \cite{chen2018isolating} and HSIC-based dependency scores \cite{gretton2005measuring} have been proposed as proxies for independence. While theoretically principled, these estimators are often computationally expensive, sensitive to hyperparameters, or lack interpretability. Our approach complements this line of work by introducing a lightweight, discrepancy-based index grounded in energy distance, which avoids density estimation altogether.

\subsection{Architectural Measures and Coupling Layers}
The role of architecture in shaping redundancy has also been studied. Work on overparameterization \cite{neyshabur2018towards} and double descent \cite{belkin2019reconciling} shows that widening networks improves optimization but can induce degenerate correlations. Work on within-network ensembles for face attributes \cite{ahmed2019within} leverages attribute correlations and grouped training to exploit inter-attribute dependencies. Coupling layers were introduced in NICE and further developed in Real NVP as efficient invertible mappings within normalizing flows \cite{dinh2016density} (see also \cite{rezende2015variational} for planar/radial flows). Inspired by this perspective, we repurpose coupling matrices not as generative components but as measurement probes: by comparing off-diagonal statistics against a normal distribution, we obtain direct evidence of redundancy accumulation across layers.

Beyond architectural design, the strategy used for exploring hyperparameters strongly influences both generalization and redundancy. Early work on random search \cite{bergstra2012random} and sequential model-based optimization \cite{bergstra2011algorithms} demonstrated that systematic exploration of configuration spaces yields more efficient discovery of performant models than grid search. These methods remain highly relevant when measuring how redundancy accumulates across architectures, since different search strategies bias the regions of parameter space that are explored.

\begin{figure*}[ht!]
    \centering
    \includegraphics[width=\textwidth]{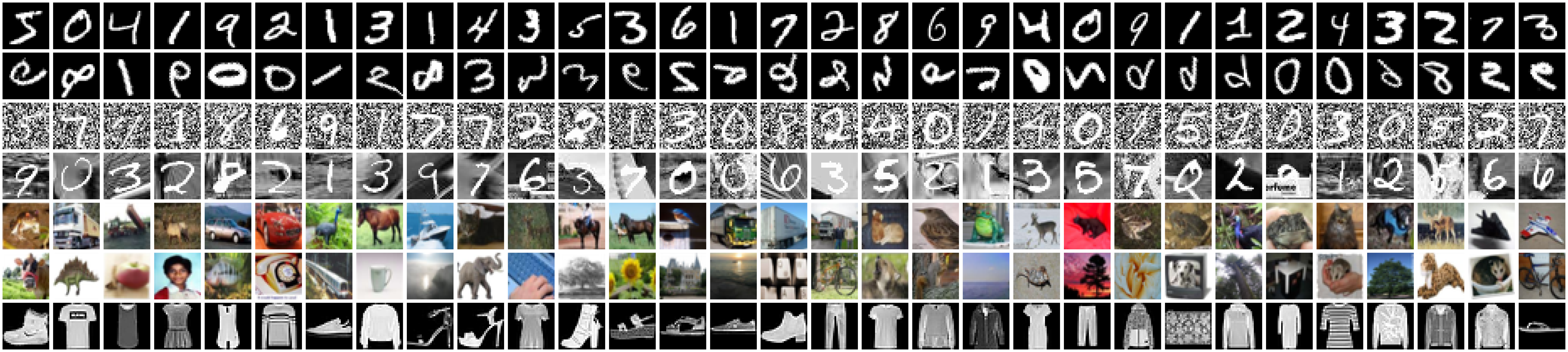}
    \caption{Sample images from benchmark datasets used in our experiments. Rows correspond to (from top to bottom): MNIST Basic, MNIST Rotated, MNIST with Random Background, MNIST with Background Images, CIFAR-10, CIFAR-100, and Fashion-MNIST.}
    \label{fig:samples}
\end{figure*}

\subsection{Evaluation Methodologies in Representation Learning}
Several benchmarks have been proposed for evaluating representation quality. Linear probing \cite{alain2018understanding} and transfer performance remain the de facto standards, but these methods are indirect and task-dependent. Unsupervised evaluation metrics such as reconstruction loss or clustering performance \cite{bengio2013representation} provide complementary insights but do not isolate redundancy as a failure mode. More recently, Larochelle et al. \cite{larochelle2007empirical} and Vincent et al. \cite{vincent2008extracting} introduced challenging variants of MNIST with structured noise and transformations, creating controlled environments for testing representational robustness. Our work builds on this tradition by systematically linking redundancy, as measured by $\rho(C)$, to both supervised and unsupervised performance across these datasets.

Complementary to these metrics, hyperparameter optimization itself can be viewed as an evaluation tool: the distribution of model outcomes across sampled configurations reflects the sensitivity of redundancy to design choices. Random search and Bayesian optimization via Tree-structured Parzen Estimators (TPE) \cite{bergstra2012random, bergstra2011algorithms} have become standard practices, enabling large-scale empirical studies that reveal systematic links between architecture, training dynamics, and redundancy indices.

\subsection{Positioning Our Contribution}
In contrast to prior methods, we propose a redundancy index that is (i) \emph{principled}, being derived from energy distance to a normal distribution model, (ii) \emph{interpretable}, since $\rho(C)$ directly quantifies deviation from an ideal disentangled regime, and (iii) \emph{general}, applying equally to discriminative and generative models. By combining theoretical grounding with broad empirical validation, our work extends the toolkit for unsupervised evaluation of learned representations and provides a bridge between information-theoretic metrics and practical measurement.

\section{Datasets}
\label{sec:datasets}

Our experiments utilize a collection of classification datasets designed to test the robustness of our method under various factors of variation, including noise, rotation, and synthetic geometric constraints. These datasets, inspired by Larochelle et al. (2007) and Vincent et al. (2008), are primarily derived from the MNIST handwritten digit dataset (LeCun et al., 1998) or generated synthetically to introduce specific challenges. Below, we describe each dataset, including its composition, preprocessing, and train/validation/test splits. All datasets and codes are publicly available \footnote{\href{https://huggingface.co/datasets/convergedmachine/Unsupervised-Evaluation-of-Latent-Space}{HuggingFace Dataset}}
\footnote{\href{https://github.com/convergedmachine/Unsupervised-Evaluation-of-Latent-Space}{GitHub Repository}}

\begin{itemize}
    \item \textbf{MNIST Basic}: This dataset is a subset of the MNIST handwritten digit dataset, consisting of 28$\times$28 pixel grayscale images of digits (0--9), each belonging to one of ten classes. To facilitate faster experimentation and highlight learning performance differences, we restructured the original train/test/validation splits. The data was randomly shuffled, resulting in 10,000 training examples, 2,000 validation examples, and 50,000 test examples. Images feature white (1.0-valued) foreground digits on a black (0.0-valued) background.
    
    \item \textbf{MNIST Background Images}: A variation of MNIST Basic, this dataset composites the white foreground digits onto 28$\times$28 natural image patches. The composition is performed by taking the pixel-wise maximum of the original MNIST image and the patch. Patches with low pixel variance were excluded to ensure visual distinctiveness. Like MNIST Basic, it includes 10 classes, with 10,000 training examples, 2,000 validation examples, and 50,000 test examples.
    
    \item \textbf{MNIST Background Random}: Similar to MNIST Basic, this dataset composites the white foreground digits onto 28$\times$28 backgrounds with pixel values drawn uniformly from (0,1). It retains the same structure as MNIST Basic, with 10 classes, 10,000 training examples, 2,000 validation examples, and 50,000 test examples.
    
    \item \textbf{MNIST Rotated}: This dataset extends MNIST Basic by applying random rotations to the images, with rotation angles sampled uniformly from $[0, 2\pi)$ radians. It includes 10,000 training examples, 2,000 validation examples, and 50,000 test examples, maintaining the 10-class structure.
    
    \item \textbf{CIFAR-10}: The CIFAR-10 dataset consists of 60,000 32$\times$32 color images drawn from 10 object classes (\emph{airplane, automobile, bird, cat, deer, dog, frog, horse, ship, truck}). Each class contains 6,000 images. Following standard splits, there are 50,000 training examples and 10,000 test examples. A validation subset of 5,000 images was sampled from the training set to facilitate hyperparameter tuning.
    
    \item \textbf{CIFAR-100}: An extension of CIFAR-10, this dataset includes 60,000 32$\times$32 color images spanning 100 fine-grained object classes, each with 600 examples. Classes are organized hierarchically into 20 superclasses, providing a richer structure for evaluation. The dataset is split into 50,000 training and 10,000 test images, with 5,000 images set aside for validation during experiments.
    
    \item \textbf{Fashion-MNIST}: Designed as a drop-in replacement for MNIST, Fashion-MNIST contains 28$\times$28 grayscale images of clothing items spanning 10 categories (\emph{T-shirt/top, trouser, pullover, dress, coat, sandal, shirt, sneaker, bag, ankle boot}). It provides 60,000 training and 10,000 test images, with 10,000 training images reserved as validation. The dataset is considered more visually challenging than MNIST Basic while maintaining the same format.
\end{itemize}

These datasets provide a diverse set of challenges, enabling comprehensive evaluation of our method's ability to handle variations in background, orientation, and geometric structure while maintaining robust classification performance (see Figure~\ref{fig:samples}).

\section{Methodology}

In this section, we develop the theoretical underpinnings of our approach to unsupervised quality measurement of latent representations. We begin by formalizing the notion of representational capacity in embedding spaces through the lens of information theory. We then introduce a coupling layer to measure inter-dimensional dependencies and propose a metric based on energy distance to measure deviation from an ideal \textit{disentangled regime}. Our framework draws on principles from multivariate statistics and discrepancy measures, providing a principled way to assess and optimize for independence in high-dimensional representations.

\subsection{Representational Capacity and Mutual Information}

The capacity of a latent embedding space to capture relevant information from the input data is a central concern in representation learning. Formally, consider an input random variable $X$ and its corresponding $d$-dimensional latent embedding $Z = (Z_1, \dots, Z_d)$. The amount of information preserved in the embedding can be quantified via the mutual information:
\[
I(X; Z) = H(Z) - H(Z \mid X),
\]
where $H(Z)$ denotes the entropy of the embedding distribution, and $H(Z \mid X)$ is the conditional entropy of $Z$ given $X$.

In deterministic encoding schemes, where $Z$ is a fixed function of $X$, the conditional entropy vanishes: $H(Z \mid X) = 0$. Consequently, maximizing $I(X; Z)$ simplifies to maximizing the entropy $H(Z)$ of the marginal distribution over embeddings. This reduction highlights entropy as a proxy for the "spread" or diversity of representations, ensuring that the latent space utilizes its full expressive power.

A key insight from information theory is that, for a fixed set of marginal distributions (or covariance constraints), the joint entropy $H(Z)$ is maximized when the components $Z_1, \dots, Z_d$ are mutually independent:
\[
H(Z) \leq \sum_{i=1}^d H(Z_i),
\]
with equality holding if and only if the $Z_i$ are independent. In the Gaussian case, this corresponds to i.i.d. components under fixed variance constraints.

Correlations among dimensions introduce redundancy: knowledge of one coordinate provides partial information about others, thereby reducing the overall joint entropy below the sum of the marginal entropies. Enforcing statistical independence ensures that each dimension contributes novel information, maximizing the informational capacity of the representation. 

The central idea is that redundancy between latent channels can be detected and quantified through the statistical structure of coupling matrices. This provides a principled criterion for selecting the most informative latent space across different hyper-parameter settings.

In the regime of ideal disentanglement:
\begin{enumerate}
    \item We introduce a coupling layer C, which reduces to the identity matrix under ideal disentanglement.
    \item The off-diagonal entries of $C$ behave as if drawn from a Gaussian noise distribution.
    \item By comparing the empirical distribution of coupling off-diagonals against this Gaussian baseline using the energy distance, we obtain a redundancy score that enables systematic evaluation of different latent spaces.
\end{enumerate}

\subsection{Coupling Matrices as Redundancy Probes}

To expose inter-dimensional interactions, we apply a learnable linear transformation to a latent representation $y \in \mathbb{R}^d$:
\[
\tilde{y} = C y, 
\qquad C \in \mathbb{R}^{d \times d},
\]
where $C$ is the coupling matrix (see Figure~\ref{fig:coupling_strategies}). 

The off-diagonal elements $\{C_{ij} : i \neq j\}$ capture dependencies between latent dimensions. In an ideally disentangled embedding, these entries should exhibit no systematic structure, reflecting only finite-sample noise. Persistent deviations instead indicate redundancy, where correlations across channels reduce the effective joint entropy of the latent space.

The coupling matrix therefore serves two complementary purposes:
\begin{enumerate}
    \item A \emph{measurement tool} for visualizing and quantifying redundancy.
    \item A \emph{practical surrogate} for estimating pairwise mutual information, since large systematic off-diagonals imply shared information between channels.
\end{enumerate}

\subsection{Ideal Noise Model for Disentanglement}

Under the null hypothesis of disentanglement, latent coordinates are uncorrelated in the population, i.e.\ $\rho_{ij}=0$ for $i\neq j$. Finite-sample estimation introduces fluctuations that can be modeled analytically. 

Let $R$ denote the sample coupling matrix computed from $n$ observations. For each off-diagonal entry $r_{ij}$, the Fisher $z$-transform
\[
z_{ij} = \sqrt{n-3}\,\mathrm{atanh}(r_{ij})
\]
is approximately distributed as $\mathcal{N}(0,1)$ under $H_0:\rho_{ij}=0$. 

This Gaussian reference distribution provides the benchmark: in a truly disentangled representation, the transformed off-diagonals should resemble standard normal noise, with no systematic deviations beyond sampling variability.

\subsection{Redundancy Index}

Let $Z$ denote the population distribution of these Fisher-transformed off-diagonals under the representation, and let $G\sim\mathcal N(0,1)$ represent the ideal noise model. The \emph{population redundancy index} is defined as the energy distance
\[
\rpop \;\triangleq\; \mathrm{ED}\!\left(Z,\,\mathcal N(0,1)\right) \;\ge 0,
\]
with $\rpop=0$ iff $Z$ matches the standard normal benchmark (perfect disentanglement).

A closed-form, finite-sample \emph{plug-in estimator} based on the empirical set $\{z_k\}_{k=1}^{\moff}$ of unique off-diagonals ($i<j$) is
\[
\rhat
= \frac{2}{\moff}\sum_{k=1}^{\moff} \mathbb{E}\!\left|z_k - G\right|
\;-\; \widehat{\mathbb{E}}\!\left|Z - Z'\right|
\;-\; \mathbb{E}\!\left|G - G'\right|,
\]
where $G,G'\!\overset{\text{i.i.d.}}{\sim}\mathcal N(0,1)$ and $Z,Z'$ are i.i.d.\ draws from the empirical distribution of $\{z_k\}$. The normal terms admit the closed forms
\[
\mathbb{E}\!\left|G - G'\right| \;=\; \frac{2}{\sqrt{\pi}}, 
\qquad
\mathbb{E}\!\left|x - G\right| \;=\; 2\phi(x) + x\bigl(2\Phi(x)-1\bigr),
\]
with $\phi,\Phi$ the standard normal pdf and cdf. The empirical self-term is given by the U-statistic
\[
\widehat{\mathbb{E}}\!\left|Z-Z'\right|
= \frac{2}{\moff(\moff-1)} \sum_{1\le i<j\le \moff} \left|z_i - z_j\right|.
\]
To ensure nonnegativity, we report $\rhatplus \triangleq \max\{0,\rhat\}$, which clips small negative values arising from finite-sample noise.

\paragraph{Redundancy Index and Estimator.}
We introduce a redundancy index $\rho(C)$, defined as the energy distance between the
distribution of Fisher-transformed off-diagonal entries $Z$ and the standard normal distribution:
\[
\rpop \;\triangleq\; \mathrm{ED}\!\left(Z,\,\mathcal N(0,1)\right).
\]
This population index is an intrinsic property of the latent representation.

In practice, $\rho$ must be estimated from finite samples. A plug-in estimator is obtained
by replacing the population distribution $Z$ with the empirical set
$\{z_k\}_{k=1}^{\moff}$:
\[
\rhat \;=\; \frac{2}{\moff}\sum_{k=1}^{\moff} \mathbb{E}\!\left|z_k - G\right|
\;-\; \widehat{\mathbb{E}}\!\left|Z - Z'\right|
\;-\; \mathbb{E}\!\left|G - G'\right|,
\]
where $G,G'\!\sim \mathcal N(0,1)$ are i.i.d.\ reference variables.
As the number of observations $n$ and latent dimension $\ndim$ grow (and thus
$\moff=\ndim(\ndim-1)/2$ increases), the estimator $\rhat$ converges to the population
index $\rpop$.

\begin{algorithm}
\caption{Energy Distance Computation}
\begin{algorithmic}[1]
\State \textbf{Input:} Coupling matrix $C \in \mathbb{R}^{d \times d}$
\State \textbf{Output:} Estimated energy distance $\widehat{\rho}(C)$

\State Extract off-diagonal entries:\\ $x \gets \{ C_{ij} : i \neq j \}, \quad n \gets d(d-1)$
\State Standardize $x$:
\State $z \gets \frac{x - \text{median}(x)}{1.4826 \cdot \text{MAD}(x)}$
\State Compute Gaussian constant: \\$\mathbb{E}|G - G'| \gets 2/\sqrt{\pi}$
\State Mixed expectation:
\For{each $z_k \in z$}
    \State $\mathbb{E}|z_k - G| \gets 2 \phi(z_k) + z_k \cdot (2\Phi(z_k) - 1)$
\EndFor 
\State Empirical self-term: \\$\widehat{\mathbb{E}}|Z - Z'| \gets \frac{2}{n(n-1)} \sum_{i<j} |z_i - z_j|$ \Comment{U-statistic}
\State Assemble ED: \\$\widehat{\rho}(C) \gets \frac{2}{n} \sum_{k=1}^n \mathbb{E}|z_k - G| - \widehat{\mathbb{E}}|Z - Z'| - \frac{2}{\sqrt{\pi}}$
\State \textbf{Return:} $\widehat{\rho}(C)$
\end{algorithmic}
\end{algorithm}

\subsection{Redundancy Parameterization}
We associate the index to a particular layer or coupling operator $C$ via the distribution of its Fisher-transformed correlation off-diagonals as above. We then write
\begin{align}
    \rpop(C) \;=\; \mathrm{ED}\!\bigl(Z(C),\,\mathcal N(0,1)\bigr) \nonumber\\
    \rhat(C),\;\rhatplus(C) \text{ for the corresponding estimators.}\nonumber
\end{align}

Interpretation:
\begin{itemize}
    \item $\rpop(C) = 0$: Off-diagonal statistics are indistinguishable from Gaussian noise (disentangled regime).
    \item $\rpop(C) > 0$: Redundancy is present; larger values indicate stronger inter-dimensional dependencies.
\end{itemize}
We use $\rhat(C)$ (or $\rhatplus(C)$) in experiments; when discussing trends informally we may write ``$\rho$'' for brevity, unless the population/estimator distinction is crucial.

\paragraph{Ablation (weight-based ED).}
As an engineering variant, we compute ED on robustly standardized off-diagonal entries of a learned coupling matrix $C$. This surrogate lacks an analytic null like Fisher $z$ but serves to probe weight-space structure. Results are included as an ablation in the Appendix \ref{app:theorem}.

\section{Experimental Evaluation}

We structure our evaluation around research questions introduced in the introduction section.  To answer these questions, we design a set of controlled experiments using multilayer perceptrons (MLPs) and autoencoders on the datasets described in Section~\ref{sec:datasets}.  We measure $\rho(C)$ on learned latent representations and compare it to task-level performance metrics such as classification accuracy or reconstruction error.

Unless otherwise stated, all reported $\rho$ values are $\widehat{\rho}$ estimates; we use $\rho$ for readability when discussing trends, and reserve the hat notation where the distinction matters (e.g., bias/variance analyses).

\begin{figure*}[htb!]
    \centering
    \includegraphics[width=\linewidth]{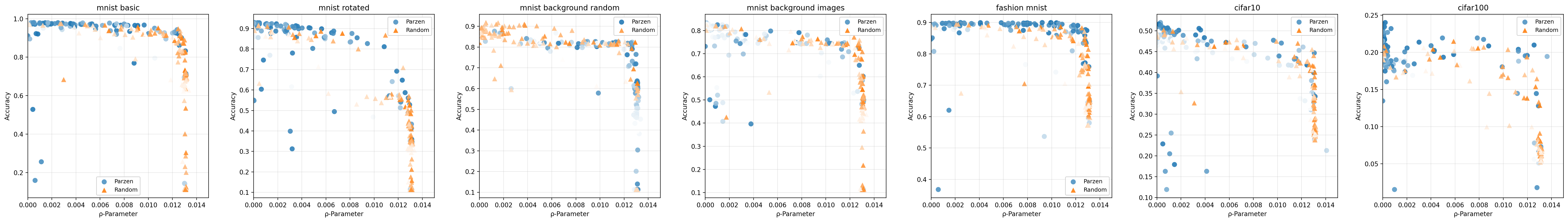}
    \caption{Accuracy versus $\rhat(C)$ for a one-layer MLP using TPE and random samplers (128 trials each). Intensity encodes network capacity.}
    \label{fig:rho_acc_1l}
\end{figure*}

\subsection*{\textbf{Experiment - 1: Correlation Between Supervised Accuracy and Redundancy Index}}

\textbf{Setup.} We evaluated 1-layer MLP classifiers across seven benchmark datasets: MNIST Basic, MNIST Rotated, MNIST with Random Background, MNIST with Background Images, Fashion-MNIST, CIFAR-10, and CIFAR-100. For each dataset, we aimed to optimize key hyperparameters that strongly influence model performance, including learning rate, batch size, weight initialization, and hidden layer dimensionality. To explore this search space, one straightforward approach is Random Search, where configurations are drawn uniformly at random. This provides a simple baseline and ensures broad coverage of possible parameter combinations. Alternatively, we employed the \textit{Tree-structured Parzen Estimator} (TPE), a sequential model-based optimization method that adaptively focuses the search on promising regions of the hyperparameter space. Unlike Random Search, TPE uses past evaluations to model the distribution of high-performing configurations and thus allocates trials more efficiently. Each method was allocated 96 independent trials, covering a wide search space of learning rate, batch size, weight initialization, and hidden dimensionality (as summarized in Table \ref{tab:hyperparams}). For every optimized model in each trial, we measured test-set classification accuracy and calculated the redundancy index $\rho(C)$ from the hidden-layer coupling matrix. To probe the relationship between model quality and representational redundancy, we plotted supervised accuracy against $\rho(C)$, separately comparing Random and TPE samplers across all datasets (see Figure \ref{fig:rho_acc_1l}).

\begin{table}[ht]
\centering
\caption{Search Space for Hyperparameter Optimization in Neural Network Models \cite{bergstra2011algorithms, bergstra2012random}}
\label{tab:hyperparams}
\resizebox{\linewidth}{!}{
\begin{tabular}{lll}
\toprule
\textbf{Hyperparameter} & \textbf{Type} & \textbf{Range / Options} \\
\midrule
Hidden units $h_1$ & Int. & 128--1024 (step 16) \\
Hidden units $h_2$ & Int. & 128--1024 (step 16), \textit{3-layer MLP or AE} \\
Hidden units $h_3$ & Int. & 128--1024 (step 16), \textit{3-layer MLP or AE} \\
Activation & Cat. & \{tanh, logistic\} \\
Initialization dist. & Cat. & \{uniform, normal\} \\
Scaling heuristic & Cat. & \{old, Glorot\} \\
Scaling multiplier & Float & 0.2--2.0 (if heuristic=old) \\
Learning rate & Log-Float & $10^{-6}$--$1.0$ \\
Anneal start & Log-Int. & 100--10000 \\
Batch size & Cat. & \{20, 100\} + \{256, 512\} \textit{only AE}\\
L2 penalty option & Cat. & \{zero, nz\} \\
L2 penalty & Log-Float & $10^{-8}$--$10^{-2}$ (if nz) \\
\bottomrule
\end{tabular}%
}
\end{table}

\textbf{Findings.}  
The scatter plots consistently demonstrate a negative correlation between redundancy ($\rho(C)$) and generalization: models with low redundancy achieve superior accuracy, whereas higher redundancy corresponds to performance collapse. This effect is most pronounced in low-dimensional image datasets (MNIST variants and Fashion-MNIST), where nearly all high-accuracy solutions cluster at $\rho(C) < 0.01$, while redundant embeddings ($\rho(C) > 0.012$) universally yield poor generalization. For CIFAR-10, lower redundancy ($\rho$) is consistently linked with higher accuracy, and the Parzen sampler more reliably discovers such configurations compared to Random search. For CIFAR-100, the single-layer MLP lacks the capacity to handle the dataset’s complexity. Most trials cluster within a narrow low-accuracy band ($\approx$ 0.15–0.22), regardless of redundancy, showing that performance here is limited by model capacity rather than redundancy.

Comparing samplers, TPE more efficiently concentrates trials in favorable low-$\rho$ regions, forming dense clusters of high-accuracy, low-redundancy models. Random Search, by contrast, distributes more evenly across the parameter space, often oversampling suboptimal, high-redundancy configurations. The general trend across all datasets highlights the dual role of $\rho(C)$ both as a measurement of representational efficiency and as a predictive marker for generalization performance. Notably, the outer boundary of the scatter clouds shows denser blue and orange points, reflecting higher network capacities; as capacity increases, both samplers tend to push against these boundaries while still respecting the redundancy–accuracy trade-off.

\begin{figure}[hb!]
    \centering
    \includegraphics[width=\linewidth]{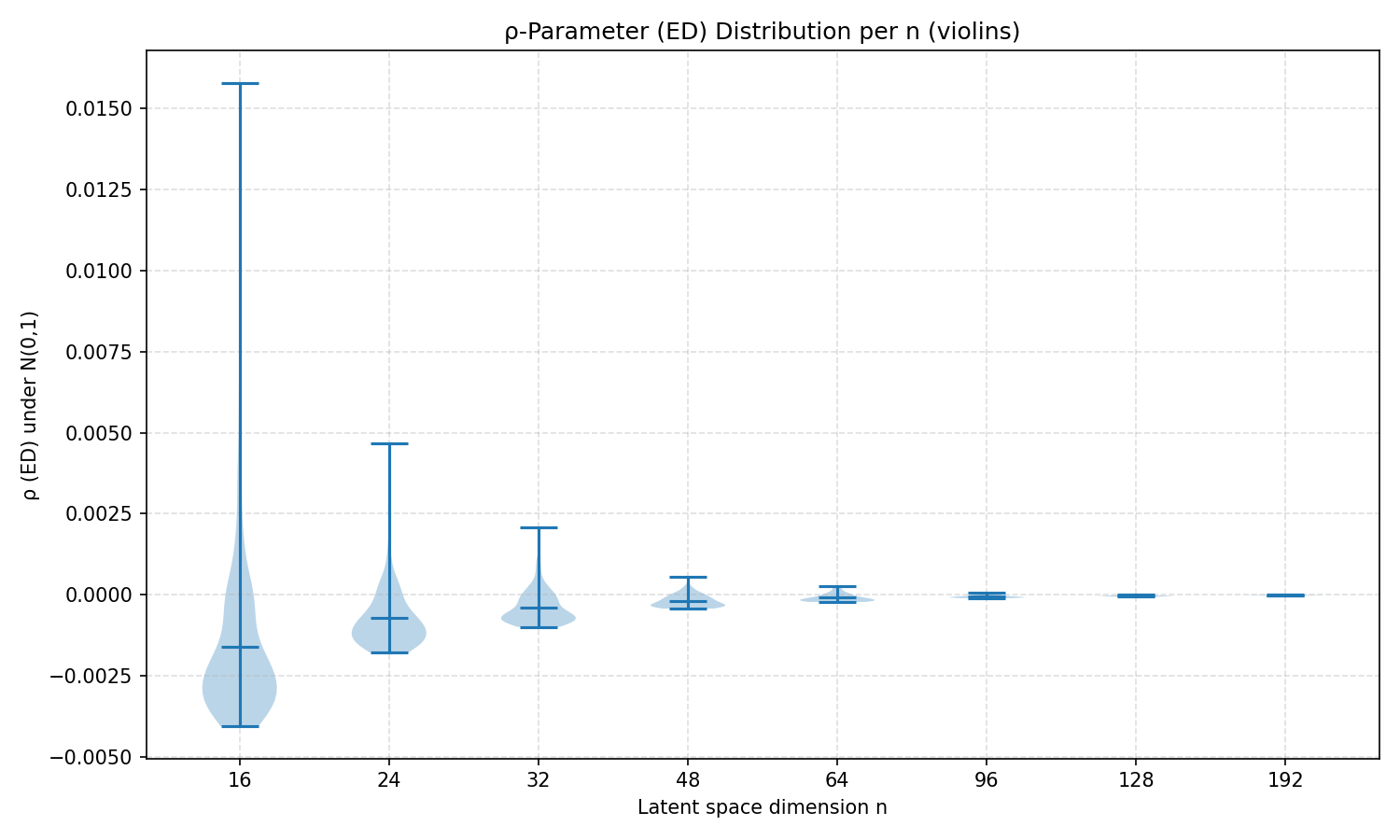}
    \caption{Violin plot of the redundancy estimator $\rhat$ across latent dimensions $n\in\{16,24,32,48,64,96,128,192\}$. The spread of $\rhat$ decreases systematically with $n$ (U-statistic variance scaling with $\moff\sim n(n-1)/2$).}
    \label{fig:rho_distributions}
\end{figure}

\subsection*{\textbf{Experiment - 2: Latent space dimensionality}}

\textbf{Setup.} We performed a systematic estimation for parameter $\rho$ based on the energy distance (ED) between empirical off-diagonal distributions of random Gaussian matrices and the standard normal distribution, for different latent dimensionalities. For each $n$, we generated symmetric and non-symmetric random matrices, extracted their off-diagonal entries, and standardized them using robust median–MAD normalization. The estimated redundancy index $\hat{\rho}$ was computed as the ED statistic for each trial, scaled to ensure a fixed total number of off-diagonal samples across $n$. Violin distributions of $\hat{\rho}$ across trials were produced as a measurement plot (see Figure \ref{fig:rho_distributions}). This setup enables quantification of small-sample bias, dependence effects from matrix symmetrization, and the stability of $\rho$ estimation across latent dimensions.

\textbf{Findings.}
The violin distributions of $\hat{\rho}$,
shown in Figure \ref{fig:samples}, reveal that both $\rhat$ bias and variance shrink systematically with increasing latent dimension size $n$. At low dimensions ($n=16,24,32$), the distributions are wide and negatively biased, while at larger $n$ the means rapidly converge toward zero with sharply reduced spread. Empirically, the dispersion of $\rhat$ decreases approximately as $1/\sqrt{\moff}$ with $\moff=\ndim(\ndim-1)/2$, consistent with classical U-statistic variance scaling. This scaling explains why the latent space stabilizes quickly once $n$ grows: the quadratic growth of off-diagonal entries dramatically increases effective sample size, yielding near-zero $\hat{\rho}$ with minimal variance by $n \geq 96$. These findings support a power-law interpretation of dimensionality effects, where redundancy estimates become progressively more reliable as the matrix dimension expands. Based on these results, we ran the next experiment with $n=96$.

\subsection*{\textbf{Experiment - 3: Dynamics of Redundancy During Training}}
\begin{figure}[htb!]
    \centering
    \includegraphics[width=\linewidth]{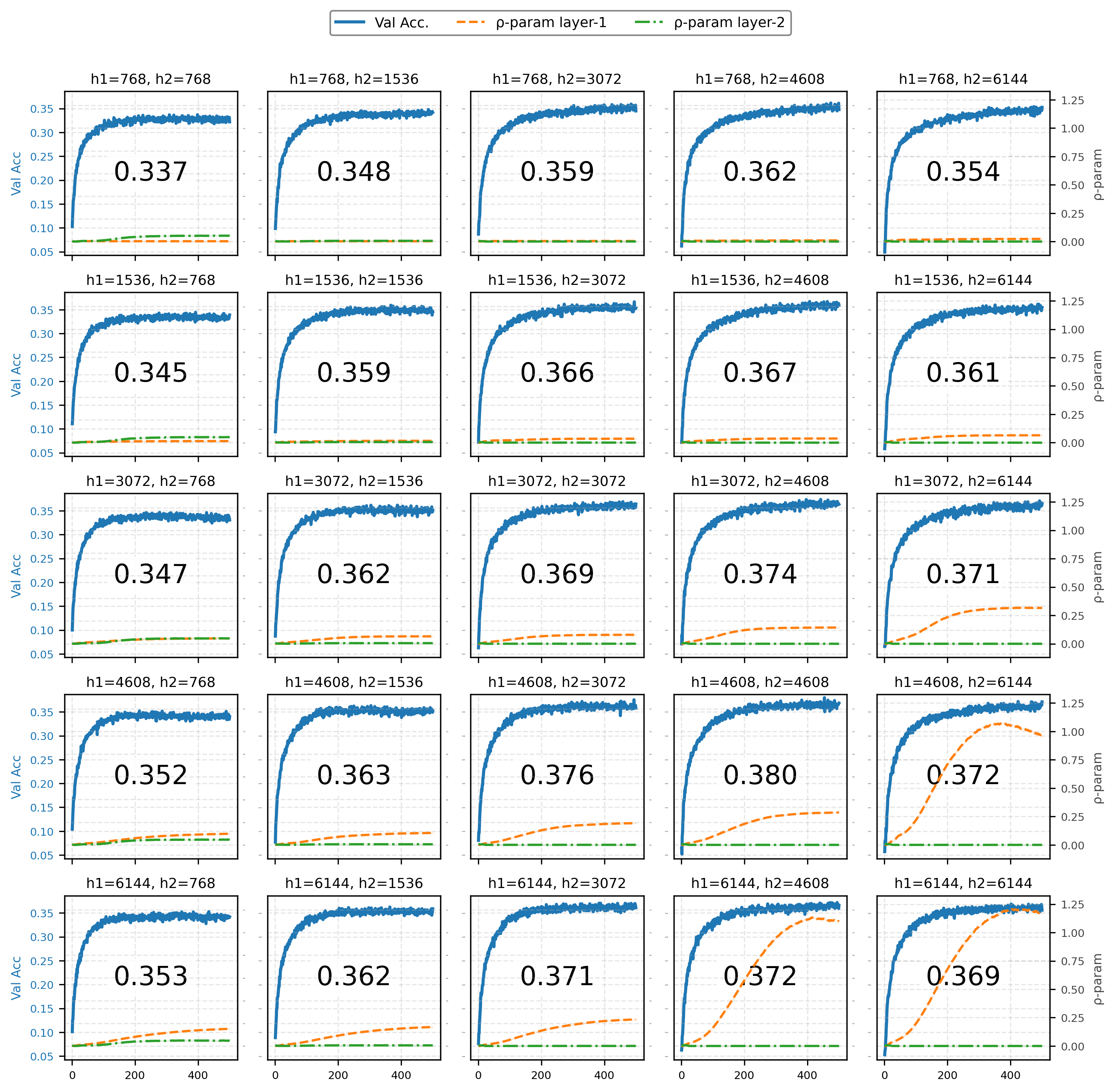}
    \caption{Validation accuracy and $\rhat$ trends for a 2-layer MLP with varying hidden unit configurations (h1, h2) on CIFAR100, optimized via grid search.}
    \label{fig:cifar100_grid_2hidden}
\end{figure}

\textbf{Setup.} 
We trained two-layer MLPs on CIFAR-100 using a comprehensive grid of hidden sizes ${768, 1536, 3072, 4608, 6144}$ for both the first ($h_1$) and second ($h_2$) hidden layers. Training employed AdamW under standard PyTorch defaults, and we monitored both the validation accuracy and the dynamics of the redundancy parameter $\rho(C)$ for each hidden layer’s coupling matrix. Each $(h_1,h_2)$ configuration was trained for 500 epochs to allow convergence. The plots report validation accuracy curves, layerwise $\rho(C)$ trajectories, and the best accuracy achieved (shown as the large centered number in each subplot; see Figure \ref{fig:cifar100_grid_2hidden}).

\textbf{Findings.}  
Validation accuracies converged rapidly within the first $\sim$75 epochs and stabilized in the range of 0.337–0.380 across all grid configurations. Performance improved systematically as hidden layer widths increased, with the best results observed for wider architectures: $(h_1,h_2)=(4608,4608)$ achieved the highest score of 0.380, while several other large settings (e.g., $(4608,3072)$, $(6144,3072)$, $(3072,4608)$) clustered closely around 0.371–0.376. In contrast, smaller or unbalanced networks (e.g., $(768,768)$, $(1536,768)$) consistently underperformed, plateauing near 0.34–0.35.

The redundancy dynamics revealed a consistent pattern across all runs. The first-layer $\rhat$ remained close to zero for small $h_1$, but when $h_1$ was very wide ($\geq$4608), $\rho(C)$ increased steadily during training, often saturating around 1.0–1.2 by the final epochs. In contrast, the second-layer $\rho(C)$ stayed comparatively low, rarely exceeding 0.25, though it exhibited mild growth when $h_2$ was wide. This asymmetry indicates that redundancy pressure shifts toward the earliest wide layer, with deeper layers contributing less to coupling accumulation.

\begin{figure*}[htb!]
    \centering
    
    \begin{subfigure}{\textwidth}
        \centering
        \includegraphics[width=\textwidth]{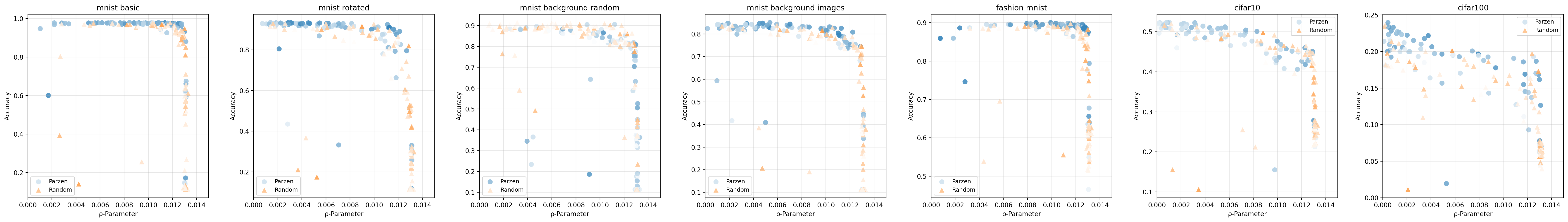}
        \caption{Performance versus $\rhat(C)$ for a three-layer MLP and an autoencoder across datasets (96 trials per sampler). Intensity encodes network capacity.}
        \label{fig:rho_acc_3l}
    \end{subfigure}
    \begin{subfigure}{\textwidth}
        \centering
        \includegraphics[width=\textwidth]{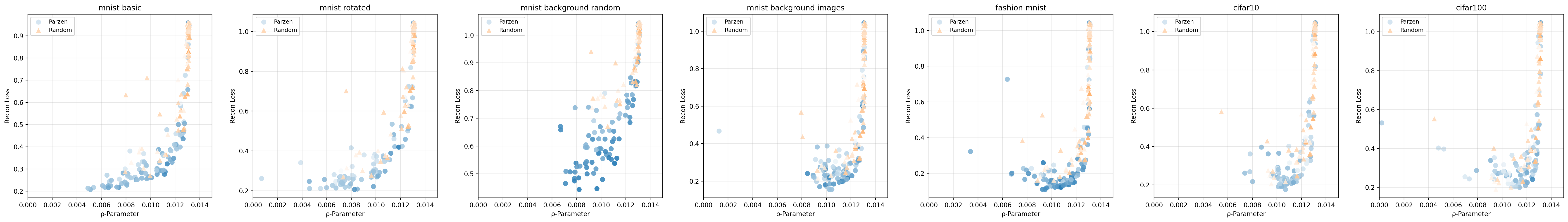}
        \caption{Reconstruction loss versus $\rhat$ distribution for an autoencoder using Parzen and random samplers (96 trials).}
        \label{fig:rho_acc_ae}
    \end{subfigure}
    
    \caption{Performance comparison of three-layer MLP and an autoencoder across diverse datasets: MNIST Basic, MNIST Rotated, MNIST Background Random, MNIST Background Images, Fashion MNIST, CIFAR10, and CIFAR100. Intensity of the color shows the network capacity.}
    \label{fig:combined_samples}
\end{figure*}

Together, these results highlight a power-law-like scaling of accuracy with hidden layer size: gains diminish beyond 4608 units, and excessive widening induces strong redundancy without proportional improvements in generalization. Optimal architectures balance width across layers, with moderate widening of both layers producing robust accuracy while avoiding the redundancy explosion observed in oversized first layers. This suggests that representational efficiency, rather than sheer parameter count, governs the trade-off between accuracy and redundancy in two-layer MLPs.

\subsection*{\textbf{Experiment - 4: Extension to Deeper Architectures}}

\textbf{Setup.}
We trained 3-layer multilayer perceptrons (MLPs) across eight benchmark datasets (MNIST Basic, MNIST Rotated, MNIST with Random Background, MNIST with Background Images, Fashion-MNIST, CIFAR-10, and CIFAR-100). For each dataset, we conducted 96 unique hyperparameter trials using two search strategies: Random Search and Tree-structured Parzen Estimator (TPE). The search space included learning rate, batch size, weight initialization, and hidden layer dimensionalities. For each trial, we recorded the supervised test accuracy and computed the redundancy index $\rho$ of the learned coupling matrix. Accuracy–$\rho$ scatter plots were generated, with blue circles representing TPE and orange triangles representing Random Search (see Figure \ref{fig:rho_acc_3l}).

\textbf{Findings.}
Across the grid of hidden layer sizes $(h_1, h_2)$ on CIFAR-100, validation accuracies stabilize in a narrow band between 0.337 and 0.380. This indicates that, for this challenging dataset, the two-hidden-layer MLPs have limited capacity, and architectural scaling only yields modest improvements. The best performance ($\approx 0.380$) is obtained for $(h_1,h_2)=(4608,4608)$, though many other large configurations achieve comparable accuracy around 0.37, suggesting diminishing returns beyond a certain capacity.

The redundancy dynamics reveal clear asymmetries between layers. The second hidden layer ($\rho$-param layer-2, green curves) remains consistently low across all configurations, implying that it contributes relatively non-redundant embeddings. By contrast, the first hidden layer ($\rho$-param layer-1, orange curves) shows a strong upward trajectory in larger-width networks, sometimes exceeding 1.0. This suggests that redundancy concentrates in the initial transformation, particularly when $h_1$ is large, potentially limiting generalization.

It is important to note that redundancy is not necessarily detrimental and need not always be minimized. In fact, some degree of redundancy can act as a form of controlled information loss, filtering out spurious or noisy variations that might otherwise hinder generalization. For example, the best-performing model $(h_1,h_2)=(4608,4608)$ achieves the highest accuracy despite a substantial increase in $\rho$ for the first layer, suggesting that selective redundancy can coexist with improved performance. At the outer edge of the scatter plots, denser clusters of blue and orange points emerge, reflecting larger network capacities; as capacity grows, both search strategies increasingly press against these limits while still adhering to the redundancy–accuracy trade-off.

\subsection*{\textbf{Experiment - 5: Redundancy in Generative Models}}

\textbf{Setup.}  
We trained a collection of autoencoders using two hyperparameter optimization strategies: Random Search and Tree-structured Parzen Estimator (TPE). For each strategy, 96 independent trials were conducted, training each model for up to 200 epochs. The hyperparameter search space included learning rate, batch size, weight initialization, and embedding dimensionality, as summarized in Table \ref{tab:hyperparams}. For every trial, we computed the mean-redundancy index $\rho_m(C)$ from the coupling matrix and measured the reconstruction loss on the test set. Scatter plots of reconstruction loss versus $\rho(C)$ were generated to visualize the dependence between representational redundancy and autoencoder performance across datasets (see Figure \ref{fig:rho_acc_ae}).

\textbf{Findings.}  
The results reveal a strong monotonic relationship between redundancy and reconstruction quality. Across all datasets, reconstruction loss remains relatively low and stable at small values of $\rho(C)$ but increases sharply once $\rho(C)$ exceed ($\approx$0.010–0.012). This behavior mirrors a power-law-like scaling: the tolerance window for redundancy narrows as dataset complexity increases, with structured-noise datasets (MNIST background random, MNIST background images) and natural images (CIFAR-10, CIFAR-100) showing the steepest degradation.

TPE consistently sampled configurations with lower $\rho(C)$, producing tighter clusters of points with reduced reconstruction loss. Random Search, in contrast, yielded a broader spread, often drifting into high-redundancy regimes associated with degraded reconstructions. The divergence between the two methods was especially evident in high-variability datasets (CIFAR-10, CIFAR-100), where TPE’s guided exploration more effectively avoided redundancy collapse. The scatter plot boundaries become denser with blue and orange points as capacity rises, showing that both samplers push toward higher-capacity limits while maintaining the redundancy–accuracy trade-off.

Overall, these findings confirm that the redundancy index $\rho(C)$ serves as a reliable predictor of reconstruction quality. Maintaining $\rho(C)$ below a level ensures stable autoencoder performance, while exceeding it results in a rapid loss of representational fidelity.

\subsection{Summary of Insights}

Our experimental results converge on several consistent insights across datasets, architectures, and training regimes:

\begin{itemize}
    \item \textbf{Latent space reliability grows with dimension.} 
    Empirically, the dispersion of $\widehat{\rho}$ decreases approximately as $1/\sqrt{\moff}$ with $\moff=\ndim(\ndim-1)/2$ (Exp.~2). 
    Owing to the quadratic growth in off-diagonal samples, the variance of the $\rho$-estimator follows a power-law decay with latent dimension $n$, yielding a natural lower bound ($n \geq 96$) for stable redundancy estimation.

    \item \textbf{Redundancy tracks generalization.} 
    Across MLPs and autoencoders, a clear trend emerges: models with $\rho(C) \lesssim 0.01$ achieve robust accuracy or low reconstruction error, while larger values typically coincide with generalization collapse. 

    \item \textbf{Width induces asymmetric redundancy.} 
    In two-layer MLPs, enlarging the first hidden layer sharply increases $\rho(C)$, often without proportional accuracy gains. Balanced designs yield stronger performance, showing that representational efficiency is constrained more by redundancy than by raw parameter count.

    \item \textbf{Redundancy is beneficial.}  
    While redundancy is often associated with degraded efficiency, it is a contructive mechanism. In fact, a measured amount of redundancy can serve as a mechanism of selective information loss, suppressing irrelevant variability and enhancing robustness. For example, the best-performing CIFAR-100 configuration $(h_1,h_2)=(4608,4608)$ achieves high accuracy despite an elevated redundancy index $\rho$ in the first layer, demonstrating that controlled redundancy can coexist with strong generalization. In this work, redundancy is quantified through the parameter $\rho(C)$, which provides a statistically grounded measure of inter-dimensional dependency.

    \item \textbf{Search strategies shape redundancy.} 
    Tree-structured Parzen Estimator (TPE) concentrates sampling in low-$\rho$ regions, consistently producing denser clusters of high-performing models compared to random search. This suggests $\rho$ can guide hyperparameter optimization.

    \item \textbf{Generative models share similar $\rho$ thresholds.} 
    Autoencoder results confirm that once $\rho(C)$ surpasses $\approx 0.010$--$0.012$, reconstruction fidelity deteriorates rapidly. This reinforces the role of $\rho$ as a general indicator of embedding quality beyond classification.

    \item \textbf{Capacity pushes against redundancy limits.}  
    Toward the boundaries of the scatter plots, denser blue and orange clusters appear, reflecting larger network capacities. As capacity increases, both sampling strategies tend to press against these limits while still maintaining the redundancy–accuracy balance.    
\end{itemize}

Together, these findings establish $\rho(C)$ as a versatile measurement of representational quality, applicable across discriminative and generative models, architectures, and optimization settings.

\begin{table*}[htb!]
\centering
\caption{Methods for quantifying divergence from normality, ranked by robustness in high-dimensional data.}
\begin{tabular}{c p{2.5cm} p{6.5cm} p{5.5cm}}
\hline
\textbf{Rank} & \textbf{Method} & \textbf{Why Robust in High-Dim?} & \textbf{Limitations} \\
\hline
\textbf{1} & Energy Distance / Energy Test & Works directly with pairwise distances; has strong consistency guarantees for multivariate distributions; widely used for high-dimensional two-sample tests. & $O(n^2)$ pairwise cost (can be approximated with subsampling). \\
\hline
\textbf{2} & Maximum Mean Discrepancy (MMD) & Kernel two-sample test; flexible kernels (e.g., Gaussian kernel adapts well to multivariate Gaussian detection). & Requires kernel bandwidth tuning; cost can be high if naive. \\
\hline
\textbf{3} & Multivariate Skewness \& Kurtosis (Mardia’s test) & Explicit multivariate generalization; captures directional asymmetry and heavy tails. & Sensitive to sample size; only captures low-order moments, misses multimodality. \\
\hline
\textbf{4} & Wasserstein-2 Distance & Conceptually strong; measures geometric discrepancy between distributions. & Computationally expensive in high-d (curse of dimensionality). \\
\hline
\end{tabular}
\label{tab:methods}
\end{table*}

\begin{table*}[htb!]
\centering
\caption{Comparison of distributional divergence metrics across different non-Gaussian distributions.}
\resizebox{\linewidth}{!}{
\begin{tabular}{lrrrrrr}
\toprule
\textbf{Distribution} & \textbf{EnergyDist.} & \textbf{MMD (RBF)} & \textbf{Mardia skew$^2$} & \textbf{Mardia excess$^2$} & \textbf{Mardia comb.} & \textbf{Wass. $W_2$} \\
\midrule
Gaussian        & -0.0003 & 0.0000 & 0.000 & 0.008 & 0.009 & 0.036 \\
$t(df=3)$       & 0.0076  & 0.0059 & 33.762 & 269141.237 & 269175.000 & 2.059 \\
Laplace (var=1) & 0.0119  & 0.0000 & 0.000 & 11.552 & 11.552 & 0.496 \\
Gaussian Mix.   & 0.0246  & 0.0188 & 0.003 & 1.727 & 1.731 & 0.320 \\
Skewed (Exp-1)  & 0.0777  & 0.0090 & 3.708 & 23.548 & 27.256 & 0.836 \\
\bottomrule
\end{tabular}
}
\label{tab:comp}
\end{table*}

\section{Discussion}

The redundancy index $\rho(C)$ provides a principled and tractable measurement for latent space quality. Unlike heuristic penalties or architectural rules of thumb, $\rho$ arises from a clear null hypothesis: in disentangled representations, coupling off-diagonals behave like Gaussian noise. Departures from this baseline admit a direct interpretation as structured redundancy among latent channels.

Several implications follow. First, $\rho$ bridges theory and practice: it is grounded in energy distance while correlating with empirical accuracy and reconstruction quality. Second, $\rho$ reveals trade-offs hidden by accuracy alone. Wider layers often inflate redundancy without proportional gains, making $\rho$ a sharper tool for architectural assessment. Third, $\rho$ is aligned with optimization dynamics: strategies like TPE implicitly discover low-$\rho$ regions, suggesting that $\rho$ could serve directly as an acquisition function in automated architecture or hyperparameter search.

Crucially, our results nuance the role of redundancy. While excess redundancy degrades generalization, controlled redundancy can act as beneficial information compression, discarding noise or irrelevant variability. This dual role highlights that $\rho$ should be interpreted as a spectrum, not a binary criterion.

At the outer edges of the scatter plots, clusters of blue and orange points become denser, reflecting the tendency of larger networks to push against the limits of capacity. Yet, even as both search strategies gravitate toward these boundaries, the redundancy–accuracy balance is largely preserved, suggesting that capacity expansion is constrained by the same trade-offs that govern smaller models.

Finally, the predictive role of $\rho$ across MLPs and autoencoders underscores its generality. Redundancy emerges as a fundamental bottleneck in representation learning, independent of model type or task, pointing toward $\rho$ as a unifying measure of embedding quality.

\section{Conclusion}

We introduced the redundancy index $\rho(C)$, derived from the energy distance between coupling off-diagonals and a normal distribution, as a measure of representational quality. Our theoretical analysis links $\rho$ to mutual information and entropy, while empirical evaluations demonstrate that it reliably predicts generalization and reconstruction outcomes across datasets, architectures, and training paradigms.

Key takeaways are: (i) $\rho$ estimation stabilizes via power-law scaling with dimension, (ii) low redundancy ($\rho \lesssim 0.01$) is a prerequisite for robust performance, though selective redundancy may be beneficial, and (iii) optimization strategies differ in their ability to explore low-$\rho$ regions, with TPE outperforming random search. Importantly, redundancy thresholds extend beyond classification to generative models, reinforcing $\rho$ as a general-purpose measurement.

In sum, the redundancy index $\rho(C)$ emerges as a principled and general measurement of latent space quality, with the potential to evolve from a post-hoc measure into a direct training signal for designing architectures that balance accuracy, efficiency, and disentanglement.

Future work may elevate $\rho$ from a measurement to a training signal, embedding redundancy-aware objectives into neural architecture design and optimization. By directly quantifying redundancy, $\rho$ offers a principled pathway toward learning representations that balance accuracy with efficiency, moving closer to disentangled, high-capacity latent spaces.

\section*{Acknowledgment}

The author would like to thank Prof. Dr. Türker Bıyıkoğlu providing thoughtful understanding and support for an early draft.

The numerical calculations reported in this paper were fully performed using the EuroHPC Joint Undertaking (EuroHPC JU) supercomputer MareNostrum 5, hosted by the Barcelona Supercomputing Center (BSC). Access to MareNostrum 5 was provided through a national access call coordinated by the Scientific and Technological Research Council of Turkey (TÜBİTAK). We gratefully acknowledge BSC, TÜBİTAK, and the EuroHPC JU for providing access to these resources and supporting this research.

\appendix
\section{Divergence from Normality Metrics}
\label{app:theorem}

This appendix provides additional detail on methods for quantifying divergence from normality.  
We present (i) a ranked comparison of methods by robustness in high-dimensional settings, and  
(ii) empirical evaluations across several non-Gaussian distributions.  

\noindent
\textbf{Table \ref{tab:methods} Interpretation.}  
Energy distance and MMD are the most reliable in high-dimensional regimes, balancing theoretical guarantees with practical usability.  
Mardia’s skewness and kurtosis are computationally lightweight but only capture moment-level deviations, making them fragile when data are multimodal or heavy-tailed.  
Wasserstein-2 distance provides strong geometric insight but becomes computationally prohibitive as dimensionality grows.  
Thus, energy distance emerges as the most practical diagnostic for large-scale representation learning tasks.

\noindent
\textbf{Table \ref{tab:comp} Interpretation.}  
For the Gaussian baseline, all methods correctly report values close to zero.  
Under heavy-tailed distributions (e.g., $t$ with $df=3$), Mardia’s kurtosis explodes, confirming its sensitivity to outliers and variance inflation.  
The Laplace distribution shows small but clear deviations in energy distance and Wasserstein metrics, while MMD fails to detect it under the chosen kernel bandwidth.  
Gaussian mixtures are well captured by energy distance and MMD, but nearly invisible to Mardia’s skewness.  
Skewed distributions (Exp-1) highlight the asymmetry detection power of Mardia’s skewness, but again energy distance provides the most stable, monotone signal across all cases.  
Overall, energy distance and Wasserstein distances consistently reflect the degree of deviation, whereas Mardia’s statistics suffer from instability, and MMD depends heavily on kernel choice. 

\bibliographystyle{IEEEtran}
\bibliography{mybibfile}

\end{document}